\PassOptionsToPackage{prologue,dvipsnames}{xcolor}

\documentclass[sigconf, screen]{acmart}

\AtBeginDocument{
  }

\setcopyright{none}
\renewcommand\footnotetextcopyrightpermission[1]{}
\usepackage{subcaption}
\usepackage{multirow}
\usepackage[dvipsnames]{xcolor}
\usepackage{makecell}
\usepackage{xcolor}   
\usepackage{pifont} 

\definecolor{cvprblue}{rgb}{0.21,0.49,0.74}
\definecolor{mygreen}{RGB}{44, 198, 117}

\newcommand{\cmark}{\textcolor{green}{\ding{51}}}
\newcommand{\xmark}{\textcolor{red}{\ding{55}}} 

\begin{document}

\title{Are Vision LLMs Road-Ready? A Comprehensive Benchmark for Safety-Critical Driving Video Understanding}

\author{
  Tong Zeng\textsuperscript{1,2},
  Longfeng Wu\textsuperscript{1},
  Liang Shi\textsuperscript{2},
  Dawei Zhou\textsuperscript{1},
  Feng Guo\textsuperscript{2,3,*}
}

\affiliation{
    \institution{
        \textsuperscript{1}Department of Computer Science, Virginia Tech;\\
        \textsuperscript{2}Virginia Tech Transportation Institute, Virginia Tech;\\
        \textsuperscript{3}Department of Statistics, Virginia Tech
    }
    \city{}
    \country{}
}

\renewcommand{\shortauthors}{Zeng et al.}

\begin{abstract}
  Vision Large Language Models (VLLMs) have demonstrated impressive capabilities in general visual tasks such as image captioning and visual question answering. However, their effectiveness in specialized, safety-critical domains like autonomous driving remains largely unexplored. Autonomous driving systems require sophisticated scene understanding in complex environments, yet existing multimodal benchmarks primarily focus on normal driving conditions, failing to adequately assess VLLMs' performance in safety-critical scenarios.To address this, we introduce \textbf{DVBench}—a pioneering benchmark designed to evaluate the performance of VLLMs in understanding safety-critical driving videos. Built around a hierarchical ability taxonomy that aligns with widely adopted frameworks for describing driving scenarios used in assessing highly automated driving systems, DVBench features 10,000 multiple-choice questions with human-annotated ground-truth answers, enabling a comprehensive evaluation of VLLMs' capabilities in perception and reasoning.
  Experiments on 14 state-of-the-art VLLMs, ranging from 0.5B to 72B parameters, reveal significant performance gaps, with no model achieving over 40\% accuracy, highlighting critical limitations in understanding complex driving scenarios. 
  To probe adaptability, we fine-tuned selected models using domain-specific data from DVBench, achieving accuracy gains ranging from 5.24 to 10.94 percentage points, with relative improvements of up to 43.59\%. This improvement underscores the necessity of targeted adaptation to bridge the gap between general-purpose vision-language models and mission-critical driving applications. DVBench establishes an essential evaluation framework and research roadmap for developing VLLMs that meet the safety and robustness requirements for real-world autonomous systems. We released the benchmark toolbox and the fine-tuned model at: \url{https://github.com/tong-zeng/DVBench.git}.
\end{abstract}



\maketitle
\thispagestyle{plain}
\pagestyle{plain}

% -------- Corresponding‑author footnote (star) ------------
\begingroup
  \renewcommand\thefootnote{\fnsymbol{footnote}} % use *, †, ‡, …
  \footnotetext[1]{Corresponding author: Feng Guo, \texttt{feng.guo@vt.edu}}
\endgroup
% ----------------------------------------------------------

\begin{figure}[h]
\centering
    \includegraphics[width=0.5\textwidth, trim=5pt 5pt 5pt 5pt, clip]{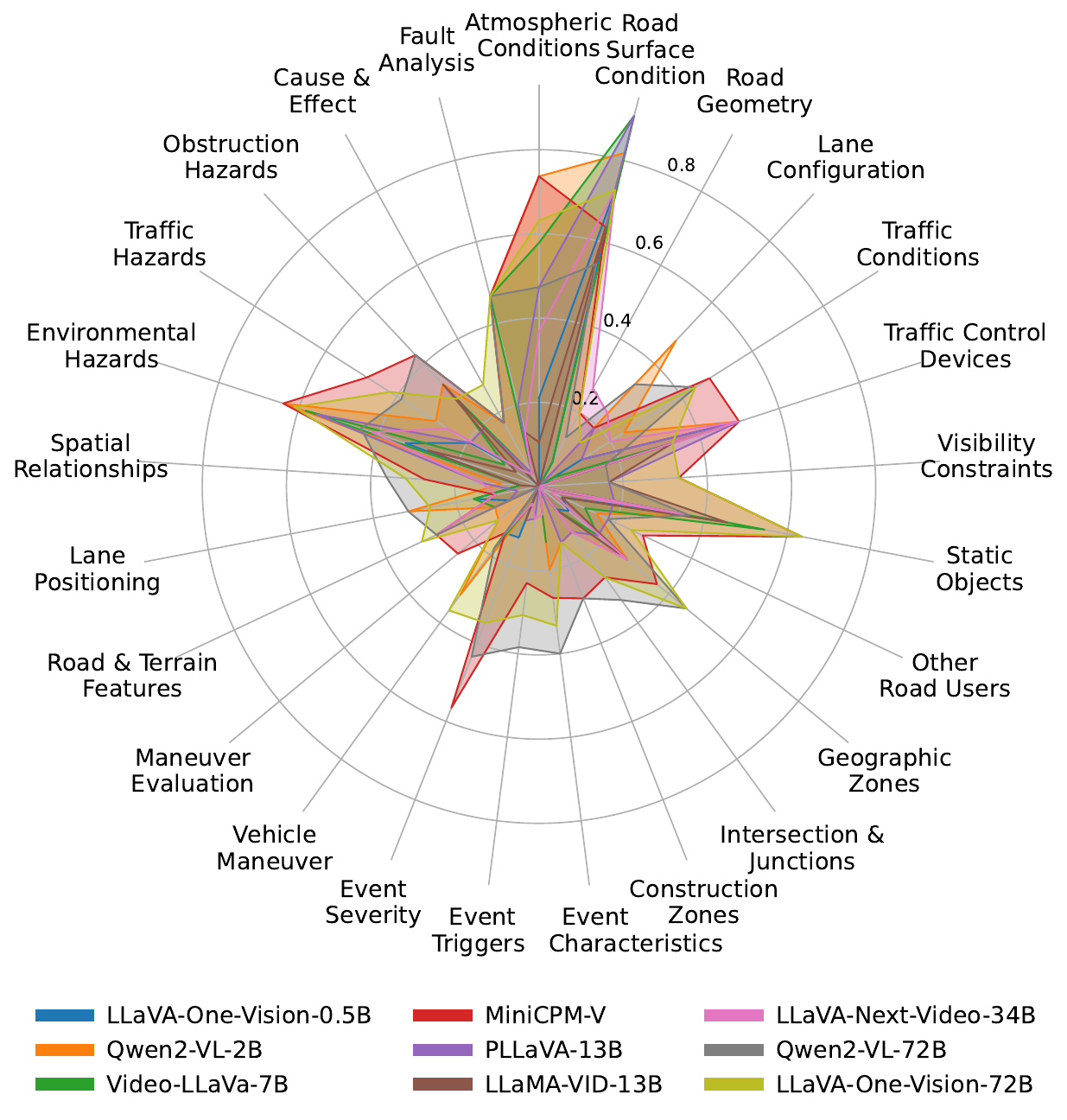}
\caption{Performance of nine representative Vision Large Language Models, ranging from 0.5B to 72B parameters, evaluated across the 25 driving video comprehension abilities defined in DVBench.}
\vspace{-5mm}
\label{fig:radar_chart} 
\end{figure}

\section{Introduction}
\label{sec:intro}

Vision Large Language Models (VLLMs)~\cite{team2023gemini, liu2024visual, alayrac2022flamingo, hurst2024gpt, zhu2023minigpt} have significantly advanced artificial intelligence by enabling machines to interpret and generate multimodal content, effectively bridging visual and textual data. Although VLLMs have demonstrated impressive performance in general-purpose applications, their potential in specialized, safety-critical domains remains underexplored. For example, Autonomous driving systems (ADS) require a comprehensive understanding of complex, dynamic environments, demanding accurate perception and interpretation to make safe, reliable decisions. This naturally raises the question: \emph{Are existing VLLMs ready for on-road deployment, especially in safety-critical scenarios?}

To answer this question, we investigate whether driving understanding generated by VLLMs reflects genuine perceptions of the environment and logical reasoning derived directly from video cues.
Recent work~\cite{mao2023gpt, xu2024drivegpt4, tian2024drivevlm, han2024dme} highlights the promise of VLLMs for ADS scene understanding. However, to ensure their effective performance in these demanding contexts, it is essential to develop comprehensive benchmarks that rigorously assess their capabilities. Evaluation remains nontrivial due to the unique challenges of ADS. 
First, (\textbf{C1. Rarity of Safety-Critical Events}) safety-critical incidents are usually rare compared to routine driving conditions, resulting in highly imbalanced datasets. Many existing benchmarks—built on popular driving datasets such as NuScenes~\cite{qian2024nuscenes}, BDD~\cite{kim2018textual}, and Waymo Open~\cite{sun2020scalability} are dominated by normal driving scenarios. This imbalance can inadvertently reward overly simplistic responses (e.g., "Going Ahead") with high accuracy~\cite{xie2025vlms}, even though safety-critical driving scenarios, which are of paramount importance for ADS, receive insufficient attention.
Second, (\textbf{C2. Lack of Temporal-Spatial Understanding in Driving Contexts}) existing benchmarks often focus on textual information or static, single-frame images, limiting their ability to capture dynamic driving scenarios. For instance, NuScenesQA~\cite{qian2024nuscenes} primarily relies on simple, one-word answers, while DriveMLLM~\cite{guo2024drivemllm} pairs front-facing camera images with linguistically diverse natural language questions. However, they largely overlook temporal-spatial perception and reasoning, essential for tracking changes over time—such as monitoring traffic signals or recognizing events leading to a crash. As a result, they fail to fully assess a model’s ability to perceive, reason, and predict in safety-critical situations.
Although general-purpose multimodal benchmarks~\cite{li2024mvbench, fu2024video, liu2025mmbench} incorporate temporal and spatial reasoning,  they are not designed for safety-critical driving scenarios. They lack evaluations specific to autonomous driving, such as interpreting traffic signs under varying conditions, analyzing pedestrian behavior, or making split-second decisions in dynamic environments.
Last, (\textbf{C3. Narrow Evaluation Scope}) some datasets prioritize specific aspects of driving. For example, DRAMA~\cite{malla2023drama} focuses on joint risk localization and captioning, while Rank2Tell~\cite{sachdeva2024rank2tell} emphasizes important object identification. 
Moreover, the construction of questions often relies on predefined templates or limited contexts, further narrowing the scope of evaluation.
Consequently, these benchmarks often lack a comprehensive understanding of driving knowledge, leading to unstable and unreliable performance in real-world applications.

To overcome these limitations, we introduce DVBench, a comprehensive benchmark designed to holistically evaluate VLLMs across a range of driving-specific capabilities (e.g., safety-critical events), including nuanced perception and reasoning. 

In particular, to address \textbf{C1}, the questions of DVBench are curated from videos of safety-critical events, such as crashes and near-crash incidents, ensuring that the evaluation covers a realistic and diverse range of challenging scenarios. To address \textbf{C2}, the DVBench questions integrate both textual descriptions and corresponding video data, ensuring a robust evaluation of VLLMs' ability to interpret and integrate multimodal information effectively. This dual-input approach enables a more nuanced assessment of models' ability to handle complex, safety-critical scenarios.
Finally, to address \textbf{C3}, DVBench is constructed around a hierarchical ability taxonomy that aligns with the requirements for scenario-based testing of ADS established by PEGASUS (Project for the Establishment of Generally Accepted quality criteria, tools and methods as well as Scenarios and Situations)~\cite{audipegasus} and driving scenario framework by the National Highway Traffic Safety Administration (NHTSA) ~\cite{thorn2018framework}. This taxonomy includes a three-level hierarchy, with 2 level-one (L1) abilities, 10 level-two (L2) abilities, and 25 level-three (L3) abilities, covering a spectrum of driving video understanding tasks from perception to reasoning. Grounding the taxonomy in high-level ADS requirements ensures that evaluated capabilities are directly relevant to real-world applications. 

Additionally, DVBench includes an automated annotation framework to generate multiple-choice questions for each task. This framework supports a curated dataset with diverse scenarios enriched by human annotations to ensure quality and reliability. To enhance the robustness of evaluations, we introduce GroupEval, a novel assessment strategy that analyzes models based on grouped capabilities and incorporates randomized option ordering. This approach mitigates position-based biases observed in large language models ~\cite{liu2024lost, kaddour2023challenges}, leading to fairer and more accurate assessments.

Our extensive experiments evaluate 14 state-of-the-art VLLMs of varying architectures and sizes, both with and without domain-specific knowledge. Results reveal significant performance gaps—no model exceeding 40\% accuracy under our GroupEval strategy, underscoring critical limitations in high-level driving perception and reasoning (see Figure~\ref{fig:radar_chart}). To mitigate these gaps, we fine-tuned selected models with DVBench data, achieving notable performance gains. These findings underscore the need for targeted adaptation and specialized domain knowledge to bridge the gap between general-purpose models and safety-critical driving applications. 
DVBench serves as a valuable benchmark for advancing VLLMs in autonomous driving by identifying model weaknesses and providing a structured evaluation framework to improve traffic safety. The main contributions of this paper are summarized as follows:

\begin{enumerate}
\item \textbf{Problem Identification}: 
We are among the first to investigate VLLMs' capabilities in perception and reasoning within safety-critical (Crash, Near-Crash) driving scenarios and systematically define the hierarchical abilities essential for evaluating the safety of autonomous driving systems in high-risk contexts.

\item \textbf{Benchmark Development}: We introduce DVBench, the first comprehensive benchmark for safety-critical driving video understanding, featuring 10,000 curated multiple-choice questions across 25 key driving-related abilities. DVBench is designed to rigorously assess perception and reasoning in dynamic driving environments.

\item \textbf{Systematic Evaluation}: We evaluate 14 state-of-the-art VLLMs, providing an in-depth analysis of their strengths and limitations in safety-critical driving scenarios. This paper establishes structured evaluation protocols and infrastructure to enable fair comparisons and guide future advancements in VLLMs for autonomous driving.

\item \textbf{Open-Source Resources}: To ensure accessibility and reproducibility, we publicly release the DVBench evaluation toolbox and fine-tuned models at: \url{https://github.com/tong-zeng/DVBench.git}.

\end{enumerate}

\section{Related Work}
In this section, we briefly introduce relevant studies on VLLMs for driving scene understanding and VLLM evaluation benchmarks.
\label{sec:RelatedWork}

\textbf{VLLMs for Driving Scene Understanding.}
The application of VLLMs in autonomous driving is an emerging area of research. DriveVLM combines VLLMs with traditional driving pipelines to address spatial reasoning challenges \cite{tian2024drivevlm}. DriveScenify leverages VLLMs to generate context-aware responses from driving videos, aiming to enhance road safety \cite{drivescenify2023multimodal}. \citet{Shoman_2024_CVPR} propose a system that integrates object detection, tracking, and language generation for detailed traffic event descriptions. \citet{jain2024semantic} enhance traffic understanding by incorporating VLLMs with multi-sensor data. DriveGPT-4 introduces an end-to-end autonomous driving system that utilizes multimodal large language models for video input processing and natural language interaction \cite{xu2024drivegpt4}. DME-Driver employs a VLLM with precise 3D scene perception to improve decision-making and control by integrating human decision logic into an autonomous driving system (ADS) \cite{han2024dme}. GPT-Driver repurposes the OpenAI GPT-3.5 model as a motion planner for ADS by framing motion planning as a language modeling task \cite{mao2023gpt}. ScVLM integrates supervised and contrastive learning to enhance the understanding and description of safety-critical events in driving scenarios \cite{shi2024scvlm}. While these studies demonstrate unique advantages in individual sub-modules such as perception, prediction, and planning, they lack comprehensive quantitative evaluations to fully assess their overall performance.

\textbf{VLLM Evaluation Benchmarks.}
Benchmarking plays a critical role in evaluating the performance of multimodal models and guiding future research directions. Popular benchmarks such as MMBench \cite{liu2025mmbench}, MV-Bench \cite{li2024mvbench}, and VideoMME \cite{fu2024video} are designed to assess multimodal models across diverse tasks using structured evaluation methodologies, including multiple-choice questions and predefined tasks, ensuring fairness and consistency in scoring. While these benchmarks provide a general framework for multimodal model evaluation, they fail to address the unique challenges inherent in autonomous driving environments.
Several driving-specific VLLM benchmarks have been proposed. Reason2Drive \cite{nie2025reason2drive} emphasizes interpretable reasoning in complex driving environments. NuScenes-QA \cite{qian2024nuscenes} provides 34,000 multi-modal visual scenes and 460,000 question-answer pairs derived from street-view data, including images and LiDAR point clouds, to assess reasoning capabilities in challenging driving scenarios. LingoQA \cite{marcu2024lingoqa} incorporates both action and scenery datasets, but its free-form answers (averaging 17.2 words) introduce significant uncertainty in evaluation outcomes. While these benchmarks address specific tasks in autonomous driving video-language learning models (VLLMs), they do not evaluate comprehensively and fail to adequately assess performance in safety-critical driving scenarios.

\section{The Proposed DVBench} 
In this section, we outline the construction of DVBench.

We first present a hierarchical ability taxonomy based on driving safety elements and autonomous system requirements, classifying essential perception and reasoning skills. We then describe data collection and human annotation processes, incorporating real-world driving data to support robust model evaluation across varied traffic scenarios and safety-critical events.

\subsection{Hierarchical Ability Taxonomy}

\begin{figure}[t] 
    \centering
    \includegraphics[width=\linewidth]{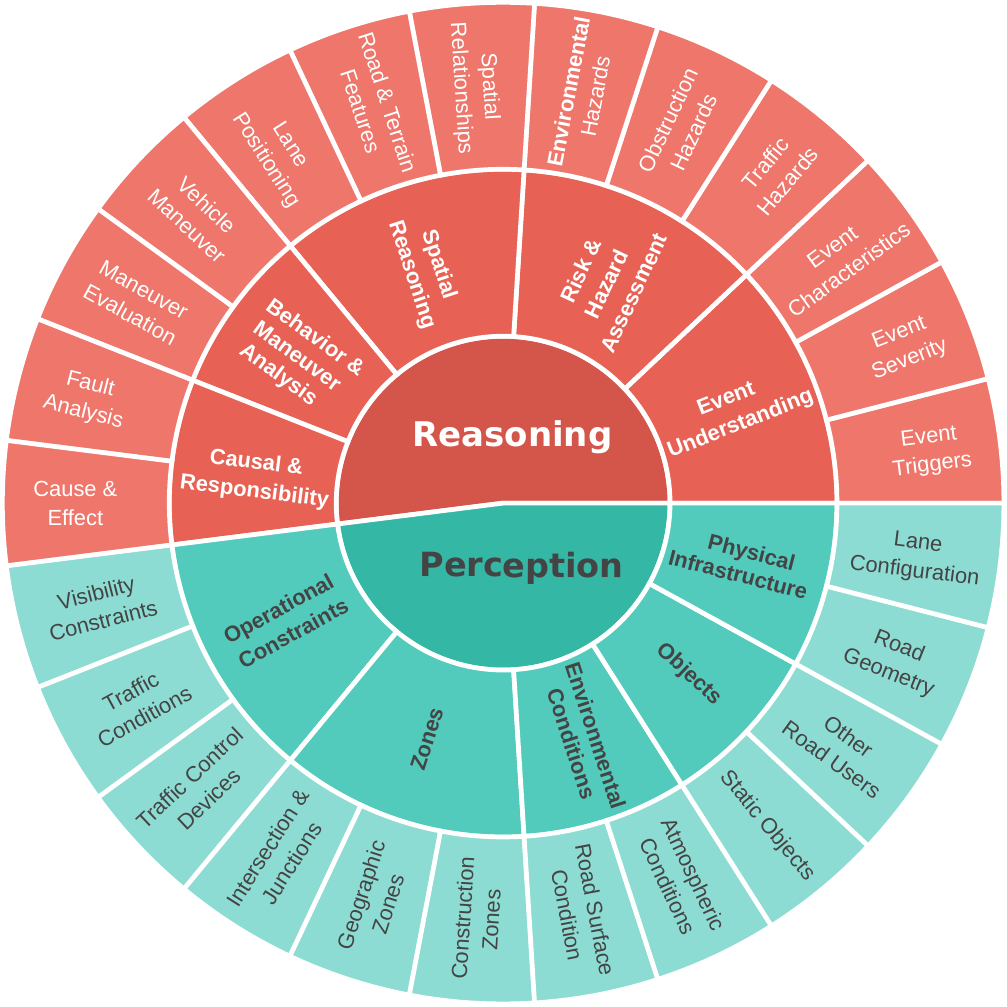}
    \caption{The DVBench Ability Hierarchy is structured across three distinct levels of capability dimensions: 2 foundational L1 abilities (perception and reasoning), 10 specialized L2 abilities for key cognitive tasks, and 25 granular L3 abilities that capture specific assessment criteria. }
    \label{fig:DVbench_hierarchy_taxonomy}
    \vspace{-5mm}
\end{figure} 

\begin{figure*}[ht] 
    \centering

\includegraphics[width=\textwidth, trim=125pt 125pt 125pt 125pt, clip]{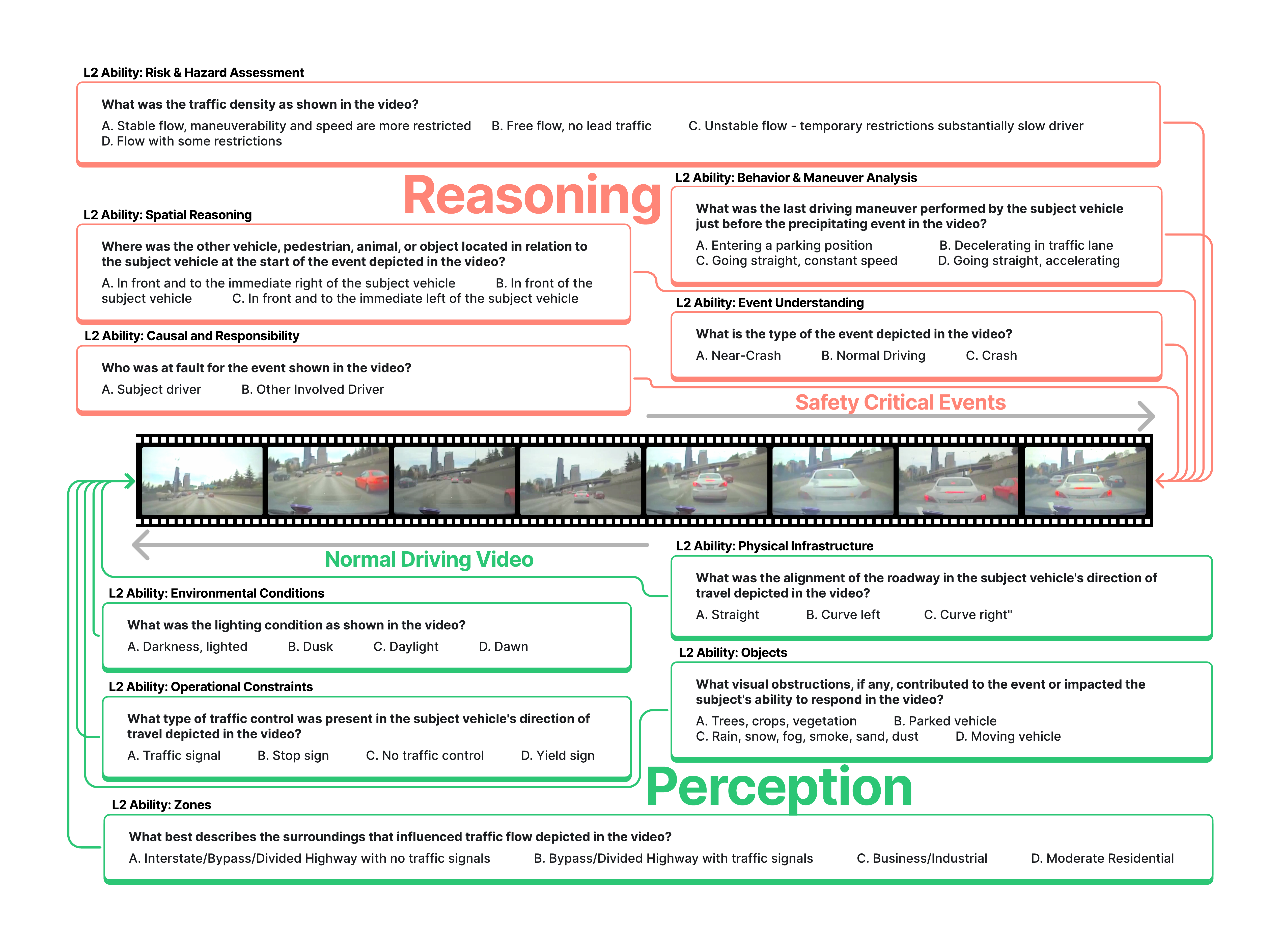}
    \caption{Examples of L2 Ability Taxonomy questions: perception and reasoning. Perception abilities focus on low-level skills such as object recognition, environmental condition parsing, and infrastructure comprehension, while reasoning abilities encompass higher-level cognitive capacities like inference, prediction, and causal analysis.}    
    \label{fig:examples of l2}
\end{figure*} 

\begin{table*}[h]
\caption{Comparison of our dataset with previous vision-language datasets related to autonomous driving (AD). "QA" stands for "Question and Answer," and "MCQ" denotes "Multiple Choice Question." A "-" indicates publicly unavailable data.}

\centering
\label{tab:benchmark_comparison}
\vspace{-2mm}
\scalebox{0.91}{
     \begin{tabular}{lccccc ccc ccc c}
    \toprule
    \multirow{2}{*}{\textbf{Benchmark}} &
    \multicolumn{2}{c}{\textbf{Tasks}} & 
    \multicolumn{3}{c}{\textbf{Driving Scenario}} & 
    \multicolumn{3}{c}{\textbf{Data Type}} & 
    \multicolumn{3}{c}{\textbf{Modality}} & 
    \multirow{2}{*}{\textbf{Amount}} \\

\cmidrule(lr){2-3} 
    \cmidrule(lr){4-6} 
    \cmidrule(lr){7-9} 
    \cmidrule(lr){10-12} 

& \textbf{Perception} & \textbf{Reasoning} & 
    \textbf{Normal} & \textbf{Crash} & \textbf{Near-Crash} & 
    \textbf{Caption} & \textbf{QA} & \textbf{MCQ} & 
    \textbf{Text} & \textbf{Image} & \textbf{Video} & \\

\midrule
    BDD-X~\cite{kim2018textual} & \cmark & \xmark  & \cmark & \xmark & \xmark & \cmark & \xmark & \xmark & \cmark & \cmark & \cmark & 26k \\
    BDD-OIA~\cite{xu2020explainable} & \cmark & \xmark  & \cmark & \xmark & \xmark & \cmark & \xmark & \xmark & \cmark & \cmark & \cmark & 35,366 \\
    Talk2car~\cite{deruyttere2019talk2car}  & \cmark & \xmark  & \cmark & \xmark & \xmark & \cmark & \xmark & \xmark & \cmark & \cmark & \xmark & 2,447 \\
    DRAMA~\cite{malla2023drama} & \cmark & \xmark  & \cmark & \xmark & \xmark & \xmark & \cmark & \xmark & \cmark & \cmark & \cmark & 17,786 \\
    NuScenesQA\cite{qian2024nuscenes}  & \cmark & \xmark  & \cmark & \xmark & \xmark & \xmark & \cmark & \xmark & \cmark & \cmark & \xmark & 83,337 \\
    Rank2Tell~\cite{sachdeva2024rank2tell}  & \cmark & \xmark  & \cmark & \xmark & \xmark & \xmark & \cmark & \xmark & \cmark & \cmark & \cmark & - \\
    DriveMLLM~\cite{guo2024drivemllm}  & \cmark & \xmark  & \cmark & \xmark & \xmark & \xmark & \cmark & \xmark & \cmark & \cmark & \xmark & 4,666 \\
    SUP-AD~\cite{tian2024drivevlm}  & \cmark & \xmark  & \cmark & \xmark & \xmark & \cmark & \xmark & \xmark & \cmark & \cmark & \xmark & - \\
    DriveLM~\cite{sima2024drivelm}  & \cmark & \cmark  & \cmark & \xmark & \xmark & \cmark & \xmark & \xmark & \cmark & \cmark & \xmark & 15,480 \\
    IDKB~\cite{lu2024can}  & \cmark & \xmark  & \cmark & \xmark & \xmark & \xmark & \cmark & \cmark & \cmark & \cmark & \xmark & 1M \\
    DriveBench~\cite{xie2025vlms}  & \cmark & \cmark  & \cmark & \xmark & \xmark & \xmark & \cmark & \cmark & \cmark & \cmark & \xmark & 20,498 \\
    \midrule
    \textbf{DVBench (Ours)}  & \cmark & \cmark  & \cmark & \cmark & \cmark & \xmark & \xmark & \cmark & \cmark & \cmark & \cmark & 10,000 \\
    \bottomrule
    \end{tabular}
}
\end{table*}

Driving scenarios encompass complex environments, road infrastructure, traffic, vehicles, and driver information.   Established frameworks like PEGASUS~\cite{audipegasus} and NHTSA~\cite{thorn2018framework}, define a six-layer structure for describing driving scenes, covering road infrastructure, temporary changes, objects, environment, and digital information. 
This layered taxonomy aligns with human perception and reasoning, categorizing L1 into "Scenario Perception" and "Scenario Reasoning,", which branch into six L2 scenario layers. Integrating this taxonomy with driving automation levels offers a structured approach to evaluating autonomous systems across the full spectrum of traffic safety considerations.

Scenario Perception focuses on gathering information about the physical road and traffic infrastructure, including temporary changes. Scenario Reasoning then analyzes dynamic objects, environmental conditions, and digital data to assess the full context.
This hierarchical taxonomy offers a structured framework for evaluating autonomous vehicle safety across diverse traffic scenarios. Figure~\ref{fig:DVbench_hierarchy_taxonomy} presents the ability taxonomy, enabling a comprehensive evaluation of VLLMs at varying abstraction levels. Detailed definitions of each fine-grained ability are provided in Appendix~\ref{sec:rationale}.

The video data in DVBench were manually collected from the Second Strategic Highway Research Program (SHRP 2)~\cite{hankey2016description}, the largest study of naturalistic driving behaviors to date.  
Over 3,300 vehicles from six sites in the US were instrumented with a data acquisition system that collected four video views (driver’s face, driver’s hands, forward roadway, rear roadway), vehicle network information (e.g., speed, brake, accelerator position), and information from additional sensors included with the data acquisition system (e.g., forward radar, accelerometers).  The driving data were collected at 10 or 15HZ.  The final SHRP2 NDS data collection consists of  4,300 years of naturalistic driving data between 2010 and 2013 with more than one million hours of continuous driving data. 

A sophisticated multi-stage process was conducted in SHRP2 NDS ~\cite{hankey2016description} to identify all safety-critical events (SCEs), including crashes and near-crashes. 
The videos of SCEs were manually analyzed and annotated through a rigorous data reduction protocol~\cite{hankey2016description}.  The annotation includes 75 variables spanning three categories: Event Variables, Driver Variables, and Environmental Variables. 

To ensure quality in data annotation, the SHRP 2 project employs a structured four-phase quality assurance and control workflow: protocol development, annotator training, active data reduction with ongoing checks, and final verification for consistency and coherence. This rigorous process maintains high data quality throughout the annotation life cycle. The annotated naturalistic driving data enriches crash analysis with detailed real-world insights.

\begin{figure*}[h]
  \centering
  \setlength{\abovecaptionskip}{2pt}
  \includegraphics[width=\textwidth, trim=120pt 120pt 120pt 120pt, clip]{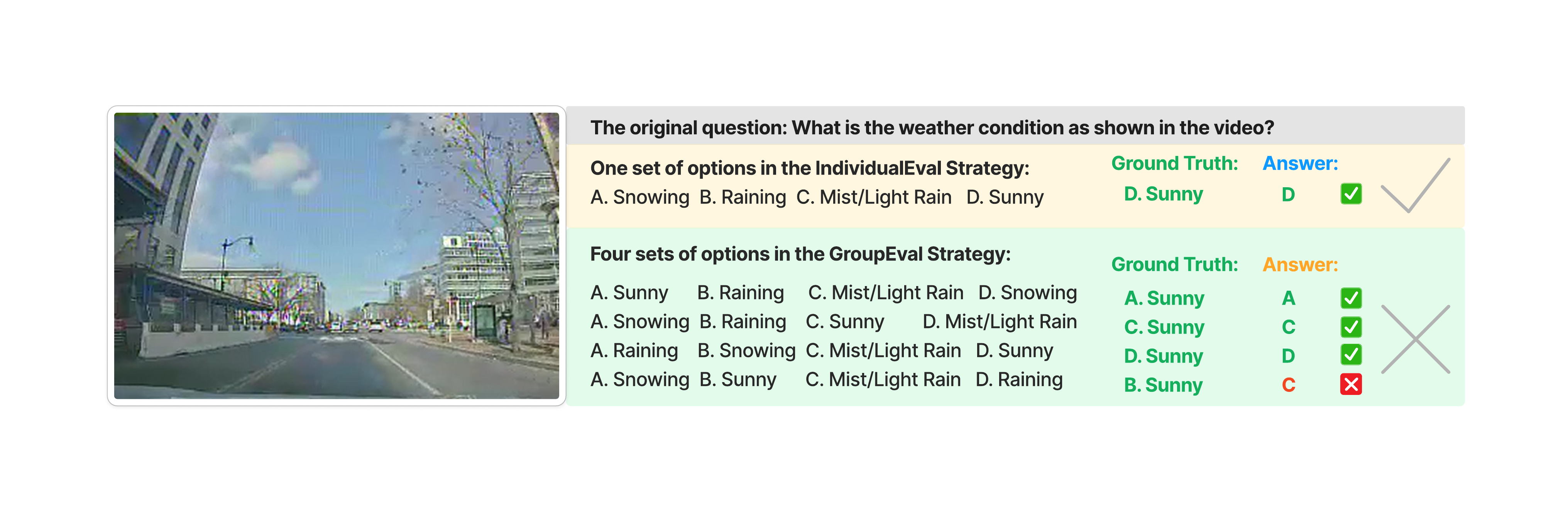}
  \caption{GroupEval Strategy. In GroupEval, each question is tested multiple times, with the correct answer's position changing each time while the other options are shuffled randomly. Here, the IndividualEval strategy deems the VLLM successful, whereas GroupEval considers it unsuccessful, as the VLLM fails to consistently identify the correct answer across different trials.}
  \label{fig:eval strategy}
  \vspace{-2mm}
\end{figure*}

\subsection{Multiple-Choice Question Bank Construction}
After obtaining the Normal Driving and Safety Critical Event videos with their annotations, we built a Multiple-Choice Question (MCQ) bank using 75 expert-annotated variables per video. This MCQ bank is designed to assess the perception and reasoning abilities of VLLMs in understanding driving scenarios. 
We have curated a total of about 10,000 (10K) MCQs with human-annotated ground truth answers spanning 25 distinct L3 abilities. A comparison between DVBench and existing benchmarks is summarized in Table~ \ref{tab:benchmark_comparison}.

Each vision-language question-answer is structured as a multiple-choice problem, defined as a quintuple $(Q_i, C_i, V_i, A_i, K_i)$, where $Q_i$ is the question, $C_i$ is a set of $n$ answer options ($c_1, c_2, ..., c_n$) with $2 \leq n \leq 4$, $V_i$ is the associated video, $A_i$ is the ground truth answer, and $K_i$ provides domain knowledge, including explanations of driving terms and test questions. In DVBench, each question-choice-answer set corresponds to a single video. 
The construction details for each element of the quintuple are introduced below.

\textbf{Video Selection and Processing.} 
DVBench consists of 4,000 video clips with a 5:4:1 ratio for Normal Driving, Crash, and Near-Crash events, extracted from 2,000 safety-critical videos. To capture short-term visual information for autonomous driving, we extracted a critical 5-second segment from each video, following the annotated event point for Crash and Near-Crash cases. The preprocessing steps include cropping timestamp watermarks and black borders, standardizing resolution and aspect ratio, resulting in 432$\times$324 clips with a 4:3 aspect ratio and a 5-second duration.

\textbf{LLM-Assisted Question Rewriting.} 
The original video data contained 75 attributes essential for understanding and evaluating autonomous driving scenarios. Using these annotations, we initially crafted a question set based on data guidelines. To better integrate domain knowledge, improve question-answer alignment, and ensure consistency, we drew inspiration from prior works \cite{li2024mvbench, zhuang2023toolqa}. In DVBench, we used prompt templates with ChatGPT o1-preview and Qwen2.5-72B models to refine questions. The process involved: 1) Filling a predefined template with manually crafted questions, annotation guidelines, candidate answers, and correct answers. 2) Feeding the prompt into the LLM to generate alternatives. 3) Manually selecting the best outputs for the question pool.

\textbf{Ground Truth and Choice Selection.} To generate multiple-choice questions, we used annotated attributes as ground truth answers and drew candidate choices from other video annotations. Question quality was ensured through a dual-filtering process: removing choices that were either frequently or rarely selected, as well as questions that all LLMs consistently answered either correctly or incorrectly. Additionally, for the multiple-choice options, we selected the $n-1$ choices from the candidate pool that were closest in length to the ground truth, thereby minimizing potential biases related to option length in model performance.

\textbf{Domain Knowledge Generation.} To improve the VLLM’s ability to accurately interpret and respond to the questions, we provided supplementary explanations encompassing traffic and driving-related terminology and context-specific scenarios. These explanations were carefully crafted based on the attribute annotation guidelines to reinforce understanding of key concepts critical for autonomous driving. We refer to this supplementary information as domain knowledge intended to bridge potential gaps in the VLLM’s comprehension of specialized driving contexts. By integrating this domain knowledge, we aim to ensure that the VLLM can more effectively handle the unique challenges posed by driving scenarios, ultimately enhancing the reliability and relevance of its responses.

\begin{figure*}[ht]
    \centering
    \setlength{\abovecaptionskip}{2pt}
    \begin{subfigure}{0.24\textwidth}
        \captionsetup{justification=centering, labelformat=empty}
        \caption{Ground-Truth Answer \\ Distribution}
        \includegraphics[width=\textwidth]{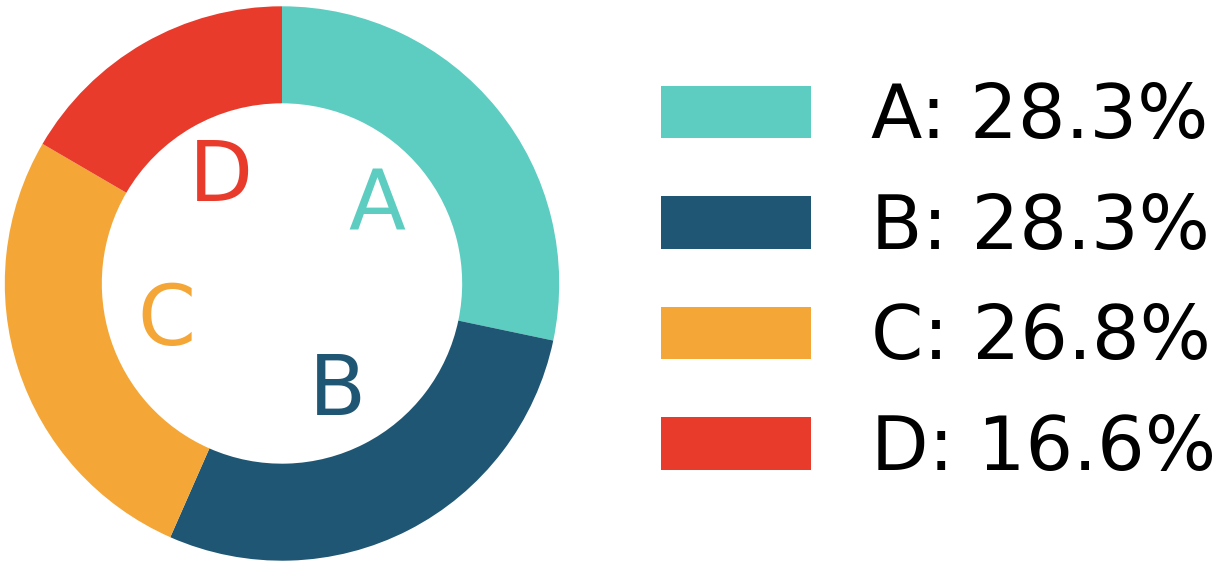}
        \label{fig:image1}
    \end{subfigure}
    \begin{subfigure}{0.24\textwidth}
        \captionsetup{justification=centering, labelformat=empty}
        \caption{Qwen2-VL-7B \\ Prediction Distribution}
        \includegraphics[width=\textwidth]{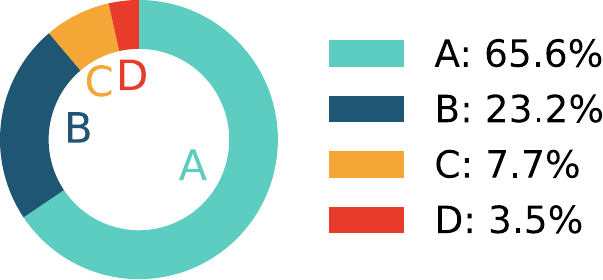}
        \label{fig:image2}
    \end{subfigure}
    \begin{subfigure}{0.24\textwidth}
       \captionsetup{justification=centering, labelformat=empty}
        \caption{PLLaVA-7B \\ Prediction Distribution}
        \includegraphics[width=\textwidth]{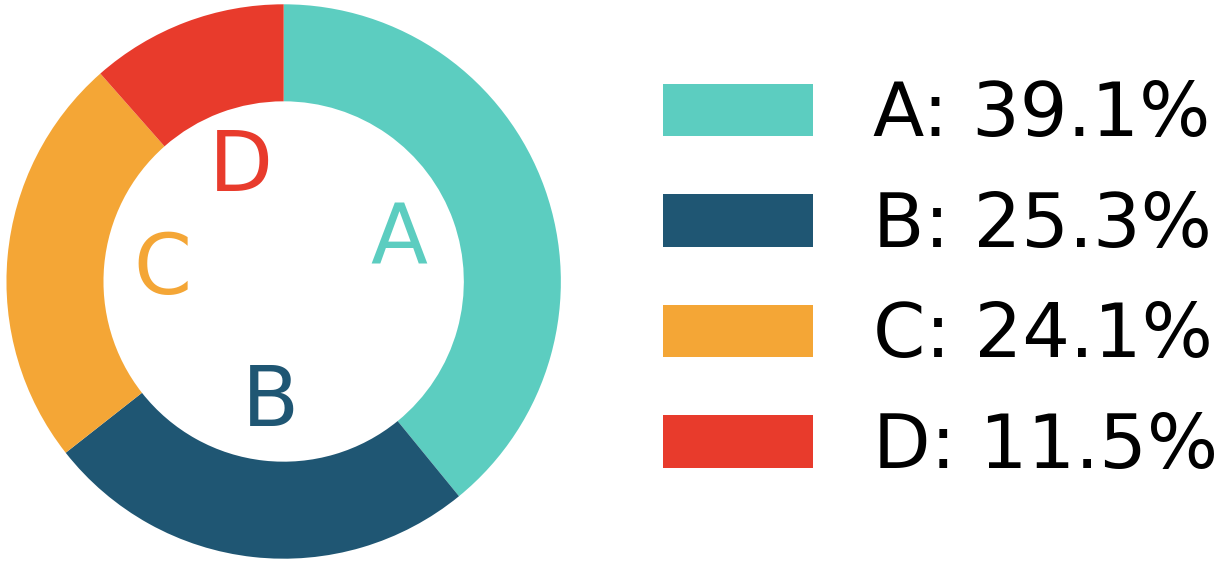}
        \label{fig:image2}
    \end{subfigure}
    \begin{subfigure}{0.24\textwidth}
        \captionsetup{justification=centering, labelformat=empty}
        \caption{LLaVA-Next-Video-7B \\ Prediction Distribution}
        \includegraphics[width=\textwidth]{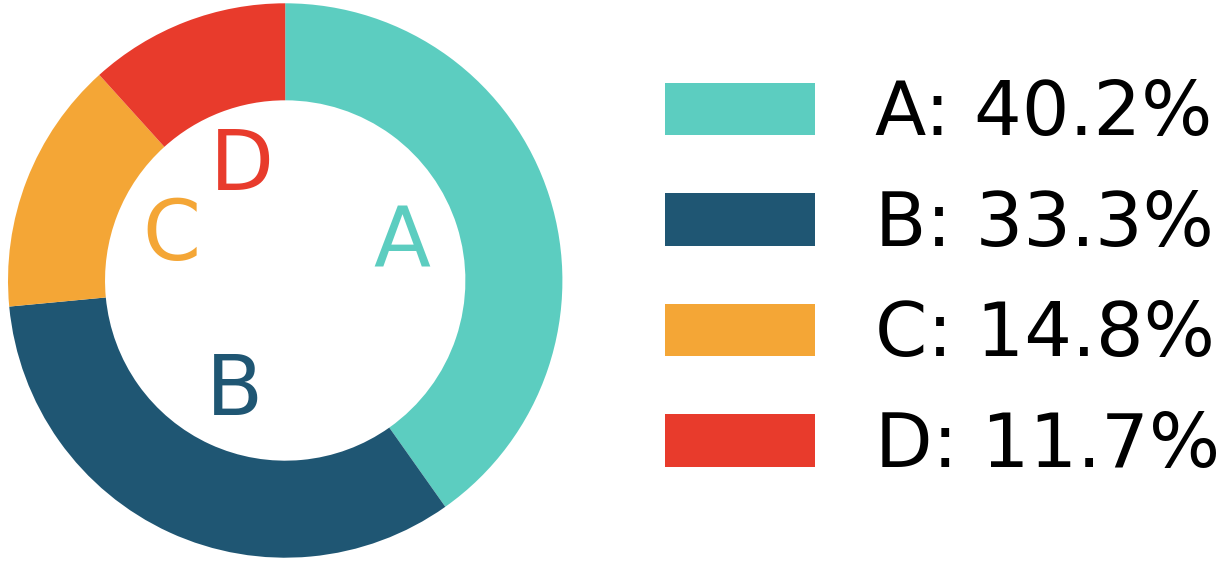}
        \label{fig:image2}
    \end{subfigure}
    \caption{Distribution of Ground-Truth Answers and Sample VLLM Predictions: In DVBench, there exist some questions that offer only 2 or 3 answer choices, leading to a slightly uneven distribution of ground-truth answers.}

\label{fig:answer distribution}
    \vspace{-2mm}
\end{figure*}

\subsection{Quality Control}
Automatically extracting question-answer pairs from video annotations does not guarantee their suitability for the benchmark.  To ensure relevance, accuracy, and consistency, we employ two key quality control mechanisms during data construction.

\textbf{Manual Checking.} We conduct manual checking for the results of LLM-assisted question rewriting and domain knowledge generation to ensure that the outputs generated by the LLM align with the expectation and to avoid LLM hallucinations. Additionally, we perform multiple rounds of sampling-checking-correction iterations for the generation of question-answer pairs. Manual cross-checking also helps identify ambiguities or inconsistencies that may arise during automated extraction, enabling necessary refinements and adjustments to enhance overall data quality.

\textbf{VLLM Majority Voting.} To further enhance question quality, we employ a majority voting strategy involving multiple state-of-the-art VLLMs. Specifically, several models generate responses for each sampled question, and those consistently yielding unanimous incorrect (or correct) answers are flagged as potentially misleading or overly simplistic. This consensus-based quality check filters out low-quality or ambiguous questions, thereby significantly improving the robustness and reliability of the benchmark dataset.

Through quality control mechanisms, we uncovered several issues impacting the evaluation process: 
\textbf{1) Lack of Visual Clarity.} Certain answers cannot be directly inferred from the video. For instance, lane counts such as "6-lane" or "7-lane" are often unclear, while conditions such as "Snowy" or "Dry" are relatively easier to identify compared to less obvious labels like "Icy." 
\textbf{2) Skewed Answer Distribution:} Some questions exhibit imbalanced answer distributions. For example, as most vehicles in the dataset did not roll over, the predominant answer to this question is "No," significantly reducing its discriminative power.
\textbf{3) Extreme Difficulty Levels.} Questions that were consistently answered incorrectly or correctly by all VLLMs were excluded, as they lack the variability required for effective benchmarking. 
\textbf{4) Similar Answer Options.} Subtly different choices, such as "Rear-end, striking" vs. "Rear-end, struck," or "Parking lot entrance/exit" vs. "Parking lot, within the boundary," is too challenging for current VLLMs to distinguish reliably. 
\textbf{5) Ambiguous Terms.} Certain Terms, such as "roadway grade," referring to road gradient or steepness, can cause confusion without further contextual clarification. 
\textbf{6) Mixed Answer Formats.} Questions that combine numerical and textual options pose additional interpretation challenges for VLLMs, complicating their ability to interpret and select the correct answer.

\section{Experiments}
In this section, we evaluate a range of VLLMs on DVBench using different evaluation strategies and the integration of domain knowledge. Additionally, we examine the impact of fine-tuning these models with DVBench data, demonstrating that leveraging domain-specific adaptation substantially enhances their performance.

\subsection{Experimental Settings}
We evaluate various \textit{VLLMs}  that capable of video understanding on DVBench, including  LLaMA-VID~\cite{li2025llama}, LLaVA-Next-Video~\cite{zhang2024llavanextvideo}, LLaVA-One-Vision~\cite{li2024llava}, MiniCPM-V (2.6) ~\cite{yao2024minicpm}, 

PLLaVA~\cite{xu2024pllava}, Video-ChatGPT~\cite{maaz2023video}, Video-LLaVa~\cite{lin2023video}, Qwen2-VL~\cite{wang2024qwen2}.

For a fair comparison, we adopt the zero-shot setting to infer DVBench questions with all VLLMs based on the same prompt template (Please refer to the Appendix for the prompt details). 
In the Appendix, we provide detailed information regarding the architecture and the parameter size for all Open-Source VLLMs evaluated in this paper, as well as additional results for more VLLMs under various settings. 

\subsection{Evaluation Strategy}
DVBench employs a multiple-choice question format, which introduces inherent evaluation challenges. Random guessing could yield approximately 25\% top-1 accuracy for 4-choice questions, potentially diminishing the observable performance differences among VLLMs. 
Furthermore, systematic biases in VLLMs complicate accurate performance assessment. For instance, many VLLMs exhibit position bias, disproportionately favoring specific answer positions, such as always selecting the first or last option~\cite{liu2024lost, chen2024premise}. This tendency is further corroborated by Figure ~\ref{fig:answer distribution}, highlighting the complexities of evaluating VLLMs fairly and reliably.

To address these challenges, we propose Group Evaluation (GroupEval), a more robust strategy. Each question is presented to VLLMs $N$ times (where $N$ equals the number of answer choices), with correct answer's position rotates systematically while other options are randomly shuffled (as illustrated in Figure~\ref{fig:eval strategy}). A VLLM passes a question only if it identifies the correct answer in all $N$ trials. This approach improves reliability while minimizing computational cost by terminating evaluation early if an incorrect prediction occurs.

\begin{table}[h]
\caption{Comparison of IndividualEval and GroupEval strategies on DVBench. VLLM accuracy is reported for both strategies, with $\Delta$ denotes the drop when switching to GroupEval.}
\centering
\label{tab: IndividualEval vs. GroupEval}
\vspace{-2mm}
\scalebox{0.98}{
 \begin{tabular}{cccc}
\toprule
\textbf{VLLMs} & \textbf{Ind. Eval} & \textbf{Grp. Eval} & \textbf{$\Delta$ } \\
\midrule
LLaMA-VID-7B~\cite{li2025llama}         & 24.9\% & 6.4\%  & {\color{red}{- 18.5\%}} \\
LLaMA-VID-13B~\cite{li2025llama}         & 18.8\% & 8.1\%  & {\color{red}{- 10.7\%}} \\
LLaVA-One-Vision-0.5B~\cite{li2024llava}  & 31.6\% & 8.1\%  & {\color{red}{- 23.5\%}} \\
Qwen2-VL-7B~\cite{wang2024qwen2}          & 36.2\% & 10.6\% & {\color{red}{- 25.6\%}} \\
LLaVA-Next-Video-7B~\cite{zhang2024llavanextvideo} & 33.0\% & 15.5\% & {\color{red}{- 17.5\%}} \\
Video-LLaVa-7B~\cite{lin2023video}        & 37.6\% & 18.0\% & {\color{red}{- 19.6\%}} \\
PLLaVA-7B~\cite{xu2024pllava}             & 34.9\% & 17.3\% & {\color{red}{- 17.6\%}} \\
LLaVA-Next-Video-34B~\cite{zhang2024llavanextvideo} & 33.1\% & 19.8\% & {\color{red}{- 13.3\%}} \\
LLaVA-One-Vision-7B~\cite{li2024llava}   & 38.6\% & 21.2\% & {\color{red}{- 17.4\%}} \\
PLLaVA-13B~\cite{xu2024pllava}           & 35.2\% & 21.6\% & {\color{red}{- 13.6\%}} \\
Qwen2-VL-72B~\cite{wang2024qwen2}          & 42.1\% & 28.6\% & {\color{red}{- 13.5\%}} \\
Qwen2-VL-2B~\cite{wang2024qwen2}           & 42.8\% & 26.9\% & {\color{red}{- 15.9\%}} \\
MiniCPM-V~\cite{yao2024minicpm}           & 46.0\% & 29.7\% & {\color{red}{- 16.3\%}} \\
LLaVA-One-Vision-72B~\cite{li2024llava}    & 45.0\% & 35.7\% & {\color{red}{- 9.3\%}} \\
\bottomrule
\end{tabular}
}
\end{table}

\subsection{IndividualEval vs GroupEval Results}
In this subsection, we compare two evaluation strategies: GroupEval (requiring consistent, correct predictions across multiple trials) and IndividualEval (single-trial inference).
Table~\ref{tab: IndividualEval vs. GroupEval} shows results from both evaluation strategies on DVBench. 
For most VLLMs, switching from IndividualEval to GroupEval results in a notable drop in accuracy. Generally, comparisons using GroupEval reveal a more substantial performance gap between different VLLMs. For example, Qwen2-VL-72B outperforms its 7B counterpart by 5.9\% in Top-1 accuracy under IndividualEval, whereas the gap widens significantly to 18\% under GroupEval. Nearly all VLLMs experience a performance drop of around 10\% to 20\% when moving from IndividualEval to the more demanding GroupEval, underscoring the latter’s challenge. For subsequent experiments, we adopt the rigorous and well-defined GroupEval as our default evaluation strategy.

\begin{table}[h]
\caption{Top-1 accuracy of VLLMs on DVBench, evaluated with (w. DK) and without (w/o. DK) domain knowledge.}
\centering
\label{tab: domain knowledge}
\vspace{-2mm}
\scalebox{0.99}{
 \begin{tabular}{cccc}
\toprule
\textbf{VLLMs} & \textbf{w/o. DK} & \textbf{w. DK} & \textbf{$\Delta$} \\
\midrule
LLaMA-VID-7B~\cite{li2025llama}         & 6.4\%  & 9.2\%  & \textbf{{\color{mygreen}{+ 2.8\%}}} \\
LLaMA-VID-13B~\cite{li2025llama}         & 8.1\%  & 8.8\%  & \textbf{{\color{mygreen}{+ 0.7\%}}} \\
LLaVA-One-Vision-0.5B~\cite{li2024llava}  & 8.1\%  & 12.7\% & \textbf{{\color{mygreen}{+ 4.6\%}}} \\
Qwen2-VL-7B~\cite{wang2024qwen2}          & 10.6\% & 25.1\% & \textbf{{\color{mygreen}{+ 14.5\%}}} \\
LLaVA-Next-Video-7B~\cite{zhang2024llavanextvideo} & 15.5\% & 15.5\% & \textbf{{\color{mygreen}{+ 0.0\%}}} \\
Video-LLaVa-7B~\cite{lin2023video}        & 18.0\% & 18.4\% & \textbf{{\color{mygreen}{+ 0.4\%}}} \\
PLLaVA-7B~\cite{xu2024pllava}             & 17.3\% & 16.6\% & {\color{red}{- 0.7\%}} \\
LLaVA-Next-Video-34B~\cite{zhang2024llavanextvideo} & 19.8\% & 18.0\% & {\color{red}{- 1.8\%}} \\
LLaVA-One-Vision-7B~\cite{li2024llava}   & 21.2\% & 21.2\% & \textbf{{\color{mygreen}{+ 0.0\%}}} \\
PLLaVA-13B~\cite{xu2024pllava}           & 21.6\% & 16.7\% & {\color{red}{- 4.9\%}} \\
Qwen2-VL-72B~\cite{wang2024qwen2}          & 28.6\% & 32.9\% & \textbf{{\color{mygreen}{+ 4.2\%}}} \\
Qwen2-VL-2B~\cite{wang2024qwen2}           & 26.9\% & 25.5\% & \textbf{{\color{mygreen}{- 1.4\%}}} \\
MiniCPM-V~\cite{yao2024minicpm}           & 29.7\% & 35.7\% & \textbf{{\color{mygreen}{+ 6.0\%}}} \\
LLaVA-One-Vision-72B~\cite{li2024llava}    & 35.7\% & 36.8\% & \textbf{{\color{mygreen}{+ 1.1\%}}} \\
\midrule
Average                                  & 19.1\% & 20.9\% & \textbf{{\color{mygreen}{+ 1.8\%}}} \\
\bottomrule
\end{tabular}
}
\end{table}

\begin{table*}[h]
\caption{Detailed performance of the VLLMs on DVBench using GroupEval (L2 abilities). The following abbreviations are used: EC (Environmental Conditions), PI (Physical Infrastructure), OC (Operational Constraints), Obj (Objects), Zone (Zones);
EU (Event Understanding), BMA (Behavior \& Maneuver Analysis), SR (Spatial Reasoning), RHA (Risk \& Hazard Assessment), CR (Causal \& Responsibility).
Additionally, the overall performance for each category (L1 abilities) is also reported.}
\centering
\label{tab:results_l2}
\vspace{-2mm}
\scalebox{0.99}{
\begin{tabular}{ccccccccccccc}
\toprule
\multirow{2}{*}{\textbf{VLLMs}} &
\multicolumn{6}{c}{\textbf{Perception}} & \multicolumn{6}{c}{\textbf{Reasoning}} \\
\cmidrule(lr){2-7} \cmidrule(lr){8-13}

& \textbf{Overall} & \textbf{EC} & \textbf{PI} & \textbf{OC} & \textbf{Obj} & \textbf{Zone}
& \textbf{Overall} & \textbf{EU} & \textbf{BMA} & \textbf{SR} & \textbf{RHA} & \textbf{CR}  \\
\midrule
LLaMA-VID-7B~\cite{li2025llama} & 12.2\% & 26.7\% & 2.8\% & 13.5\% & 11.4\% & 9.1\% 
& 10.4\% & 2.0\% & 3.3\% & 7.5\% & 19.3\% & 11.4\% \\
LLaMA-VID-13B~\cite{li2025llama} & 12.8\% & 30.0\% & 0.0\% & 8.1\% & 15.9\% & 12.1\% 
& 9.6\% & 3.9\% & 0.0\% & 6.0\% & 21.6\% & 4.5\% \\
LLaVA-One-Vision-0.5B~\cite{li2024llava} & 15.0\% & 40.0\% & 2.8\% & 16.2\% & 13.6\% & 6.1\% & 11.1\% & 5.9\% & 10.0\% & 9.0\% & 21.6\% & 0.0\% \\
Qwen2-VL-7B~\cite{wang2024qwen2} & 25.6\% & 43.3\% & 2.8\% & 27.0\% & 27.3\% & 30.3\% & 
27.1\% & 33.3\% & 13.3\% & 17.9\% & 33.0\% & 31.8\% \\
LLaVA-Next-Video-7B~\cite{zhang2024llavanextvideo} & 18.3\% & 43.3\% & 2.8\% & 16.2\% & 18.2\% & 15.2\% 
& 16.1\% & 7.8\% & 13.3\% & 10.4\% & 27.3\% & 13.6\% \\
Video-LLaVa-7B~\cite{lin2023video} & 21.1\% & 70.0\% & 2.8\% & 8.1\% & 22.7\% & 9.1\% 
& 18.6\% & 5.9\% & 13.3\% & 10.4\% & 34.1\% & 18.2\% \\
PLLaVA-7B~\cite{xu2024pllava} & 20.0\% & 56.7\% & 2.8\% & 10.8\% & 18.2\% & 18.2\% 
& 17.9\% & 3.9\% & 10.0\% & 10.4\% & 34.1\% & 18.2\% \\
LLaVA-Next-Video-34B~\cite{zhang2024llavanextvideo} & 23.3\% & 50.0\% & 25.0\% & 21.6\% & 11.4\% & 15.2\% 
& 16.1\% & 5.9\% & 10.0\% & 17.9\% & 27.3\% & 6.8\% \\
LLaVA-One-Vision-7B~\cite{li2024llava} & 28.3\% & 70.0\% & 19.4\% & 21.6\% & 20.5\% & 18.2\% 
& 20.0\% & 3.9\% & 20.0\% & 16.4\% & 37.5\% & 9.1\% \\
PLLaVA-13B~\cite{xu2024pllava} & 23.9\% & 63.3\% & 11.1\% & 18.9\% & 18.2\% & 15.2\% 
& 15.4\% & 5.9\% & 0.0\% & 9.0\% & 34.1\% & 9.1\% \\
Qwen2-VL-72B~\cite{wang2024qwen2} & 32.8\% & 50.0\% & 25.0\% & 35.1\% & 22.7\% & 36.4\% & 
33.9\% & 41.2\% & 13.3\% & 31.3\% & 42.0\% & 27.3\% \\
Qwen2-VL-2B~\cite{wang2024qwen2} & 31.7\% & 76.7\% & 27.8\% & 24.3\% & 20.5\% & 18.2\% & 
26.4\% & 7.8\% & 26.7\% & 16.4\% & 44.3\% & 27.3\% \\
MiniCPM-V~\cite{yao2024minicpm} & 39.4\% & 70.0\% & 19.4\% & 45.9\% & 36.4\% & 30.3\% 
& 35.4\% & 39.2\% & 16.7\% & 22.4\% & 53.4\% & 27.3\% \\
LLaVA-One-Vision-72B~\cite{li2024llava} & 36.7\% & 66.7\% & 16.7\% & 40.5\% & 34.1\% & 30.3\% 
& 36.4\% & 33.3\% & 30.0\% & 29.9\% & 46.6\% & 34.1\% \\
\bottomrule
\end{tabular}
}
\vspace{-2mm}
\end{table*}

\subsection{Domain Knowledge Matters}
\label{domain knowledge matters}

In this subsection, we explore whether domain-specific knowledge can boost VLLM performance. Autonomous driving, in particular, involves concepts that may be unfamiliar to those outside the field, which can present a challenge for these models. To address this, we add clarifying information after each question to provide the necessary context.
For instance, consider the question: \textit{“What was the relation of the subject vehicle to the junction depicted in the video?”} While straightforward to experts, this phrasing may be unclear to individuals (or models) without driving experience. By providing domain knowledge, such as: \textit{``Relation to Junction: The spatial (rather than causal) relationship of the subject vehicle to a junction at the beginning of the Precipitating Event. A junction is ... typically with differing travel speeds or directions."}, we make the question more accessible. This contextual information can help general-purpose VLLMs better understand the question, potentially improving their accuracy and effectiveness in such domain-specific tasks.

In Table~\ref{tab: domain knowledge}, we report the top-1 accuracy of VLLMs on DVBench with and without domain knowledge. 
Most models show a significant accuracy boost with domain knowledge, highlighting the value of contextual understanding. 
However, certain models, such as PLLaVA-13B~\cite{xu2024pllava}, experience a slight decline in domain knowledge. We believe this may stem from factors such as misinterpretation of domain-specific terminology or limitations in the model’s ability to generalize under the added complexity of domain-focused information. This suggests that while domain knowledge enhances understanding, its integration must be carefully calibrated to avoid adverse effects on certain models.

\subsection{Main Results}

\textbf{Overall Evaluation.} We evaluated various VLLMs on DVBench across two key dimensions: Perception and Reasoning, each with distinct sub-categories aligned with autonomous driving evaluation requirements. L2 ability evaluation results are shown in Table~\ref{tab:results_l2}, while more fine-grained L3 results appear in Figure~\ref{fig:radar_chart}.
The LLAVA-One-Vision-72B and MiniCPM-V achieved the highest overall scores, with perception accuracies of 39.4\% and 36.7\%, and reasoning accuracies of 35.4\% and 36.4\%, respectively. Notably, no model surpassed 40\% overall accuracy, highlighting a significant gap between current VLLM capabilities and the stringent requirements for real-world autonomous driving. In such applications, even minor errors pose serious safety risks, underscoring the need for extensive domain-specific adaptation and training.

\noindent\textbf{Perception Outperforms Reasoning.}
In the Perception tasks, Qwen2-VL-72B achieved the highest score in Environmental Conditions (EC) with 76.7\%, followed by Video-LLAVA-7B and LLAVA-One-Vision-7B at 70.0\%. For Physical Infrastructure (PI), only Qwen2-VL-72B and LLAVA-One-Vision-72B showed relatively strong performance. In Object (Obj) detection, MiniCPM-V led with 36.4\%, closed followed by LLAVA-One-Vision-72B at 34.1\%.
For Reasoning tasks, LLAVA-One-Vision-72B scored highest in Risk \& Hazard Assessment (RHA) with 46.6\% and achieved 34.1\% in Causal \& Responsibility (CR), indicating some capacity for hazard assessment and causality attribution. However, overall reasoning performance remains limited.
These results show that VLLMs excel in perception but struggle with reasoning—an essential capability for autonomous driving. The gap in reasoning highlights significant challenges in real-world deployment, where robust, context-aware decision-making is critical for safety and reliability.

\noindent\textbf{Foundation Models: A Key but Not Sole Factor in Success} 
Contrary to the common belief that larger models are always superior, DVBench results indicate that this is not necessarily the case. For instance, MiniCPM-V (8B parameters) outperforms the larger LLaVA-One-Vision-72B in perception tasks and performs comparably in reasoning tasks.
We observed a trend that models using Qwen2 as the language backbone and SigLip~\cite{zhai2023sigmoid} as the vision encoder tend to achieve strong performance. While this suggests that foundation models play a crucial role in success, they are not the sole determining factor, as model performance is influenced by a combination of architecture, training data, and optimization strategies. This highlights the need for a holistic approach to model design and training for autonomous driving applications.

\subsection{Impact of Fine-Tuning on VLLMs}

Thus far, our evaluation has focused on general-purpose VLLMs in safety-critical driving videos. While these models serve as foundation models that underpin downstream tasks, they are not optimized for autonomous driving. To assess their adaptability, we fine-tune VLLMs using safety-critical driving videos, measuring their post-fine-tuning performance and the improvements achieved.

\noindent\textbf{Fine-Tuning Setup.} To evaluate the impact of fine-tuning, we selected the Qwen2-VL series as our backbone model, as it covers a wide range of parameter scales and demonstrates strong baseline performance among available VLLMs. Specifically, we fine-tuned Qwen2-VL-2B and Qwen2-VL-7B on 2,880 human-annotated question-answer pairs from DVBench, ensuring alignment with real-world driving scenarios.

\noindent\textbf{Performance Gains from Fine-Tuning.} Table~\ref{tab:accuracy_comparison} presents the accuracy comparison between fine-tuned and non-fine-tuned Qwen2-VL models. We observe notable improvements in accuracy, particularly for the Qwen2-VL-7B model, which exhibited a +10.94 percent points increase after fine-tuning, suggesting that larger models benefit significantly from domain adaptation. Similarly, the Qwen2-VL-2B model showed a +5.24 percent points improvement, while Qwen2-VL-72B was excluded from fine-tuning due to computational constraints.

\noindent\textbf{Analysis and Implications.} The results demonstrate that task-specific fine-tuning significantly enhances VLLMs' comprehension of driving scenarios, reinforcing the necessity of domain-specific adaptation to bridge the gap between general-purpose visual reasoning and high-stakes decision-making in autonomous driving. While VLLMs exhibit strong baseline capabilities, our findings highlight that pre-trained models alone are not sufficiently reliable for safety-critical applications. Effective fine-tuning and optimization are essential to ensure deployment-ready accuracy, making adaptation strategies a crucial step toward improving VLLMs' viability in real-world autonomous driving systems.

\begin{table}[h]
\caption{Accuracy comparison of various VLLM models with and without fine-tuning.}
\centering
\vspace{-2mm}
\label{tab:accuracy_comparison}
    \begin{tabular}{l p{0.17\linewidth} p{0.17\linewidth} >{\centering\arraybackslash}p{0.17\linewidth}}
        \toprule
        \textbf{VLLM Model} & \textbf{w/o. FT} & \textbf{w. FT} & \textbf{$\Delta$} \\
        \midrule
        Qwen2-VL-2B    & 25.5\%  & \textbf{30.74\%} & \textbf{{\color{mygreen}{+5.24\%}}} \\
        Qwen2-VL-7B    & 25.1\%  & \textbf{36.04\%} & \textbf{{\color{mygreen}{+10.94\%}}} \\
        \bottomrule
    \end{tabular}
\vspace{-5mm}
\end{table}

\section{Conclusion and Future Work}
We introduced DVBench, the first benchmark specifically designed to evaluate VLLMs in safety-critical driving video understanding. Built around a hierarchical ability taxonomy rooted in widely adopted frameworks for describing driving scenarios, DVBench structures its evaluation into three levels across 25 essential driving safety abilities. Our evaluation of 14 state-of-the-art VLLMs, ranging from 0.5B to 72B parameters, revealed substantial performance gaps, with no model surpassing 40\% accuracy under our GroupEval strategy. Fine-tuning with domain-specific data achieves accuracy gains of up to 10.94 percentage points, underscoring the necessity of targeted adaptation to bridge the gap between general-purpose VLLMs and mission-critical driving applications.

Future work on DVBench will focus on expanding its evaluation scope and improving robustness. One key direction is introducing new question types, such as True/False and open-ended questions, to better assess VLLMs’ reasoning. Additionally, we will develop a VLLM arena with public leaderboards to enable continuous benchmarking and foster research competition. Refining the dataset with more rare and long-tail driving events will further enhance its reliability. By evolving DVBench, we aim to provide a stronger benchmark for advancing VLLMs in autonomous driving.

\bibliographystyle{ACM-Reference-Format}
\bibliography{main.bib}

\appendix
\clearpage
\setcounter{page}{1}

\setcounter{section}{0}

\section{Details of Hierarchical Ability Taxonomy }
\label{sec:rationale}

The DVBench Ability Taxonomy consists of two fundamental components that capture the multifaceted nature of visual-linguistic understanding: perception and reasoning abilities. Perception abilities encompass foundational skills such as object recognition, environmental condition parsing, and infrastructure comprehension. Reasoning abilities, on the other hand, involve sophisticated cognitive processes including inference, prediction, and causal analysis. By organizing the taxonomy along these two dimensions, DVBench offers a comprehensive framework for evaluating how effectively Vision-Language Large Models interpret and synthesize complex visual-textual scenarios. The following subsections provide detailed definitions for each L3 ability within this taxonomy.

\begin{table*}[h]
\caption{The vision encoder and language model details of the evaluated VLLMs.}
\centering
\label{tab:details_of_VLLMs}
\resizebox{0.99\textwidth}{!}{
\begin{tabular}{l|cccc}
\toprule
\textbf{VLLMs}  & \textbf{Vision Encoder}  & \textbf{Vision Encoder Size}  & \textbf{Language Model} &  \textbf{LLM Size} \\
\midrule
LLaMA-VID-7B~\cite{li2025llama} & EVA-G & 1.1B & Vicuna & 7B   \\
LLaMA-VID-13B~\cite{li2025llama} & EVA-G & 1.1B  & Vicuna & 13B   \\
LLaVA-Next-Video-7B~\cite{zhang2024llavanextvideo} & SigLIP  & 400M & Vicuna-v1.5 &  7B    \\
LLaVA-Next-Video-34B~\cite{zhang2024llavanextvideo} & SigLIP  & 400M & Vicuna-v1.5 &  34B    \\
LLaVA-One-Vision-5B~\cite{li2024llava} & SigLIP   & 400M  & Qwen-2 &  5B    \\
LLaVA-One-Vision-7B~\cite{li2024llava} &  SigLIP  & 400M & Qwen-2 & 7B    \\
LLaVA-One-Vision-72B~\cite{li2024llava} & SigLIP  & 400M & Qwen-2 &  72B    \\
MiniCPM-V~\cite{yao2024minicpm} & SigLIP  &  400M &  Qwen2 &  7B \\
PLLaVA-7B~\cite{xu2024pllava} & CLIP ViT-L  & 428M  & Vicuna-v1.5 &  7B  \\
PLLaVA-13~\cite{xu2024pllava} & CLIP ViT-L  & 428M  & Vicuna-v1.5 &  13B  \\
Qwen2-VL-2B~\cite{wang2024qwen2} & RoPE-2D+ViT & 675M & Qwen2  &  1.5B\\
Qwen2-VL-7B~\cite{wang2024qwen2} &  RoPE-2D+ViT  & 675M  & Qwen2  &  7.6B\\
Qwen2-VL-72~\cite{wang2024qwen2} & RoPE-2D+ViT  &  675M & Qwen2  & 72B\\
Video-LLaVa-7B~\cite{lin2023video} & OpenCLIP-L/14 & 428M & Vicuna-v1.5 &  7B \\
\bottomrule
\end{tabular}
}
\end{table*}

\vspace{10mm}

\begin{table*}[h]
\caption{Detailed performance of the VLLMs on DVBench using GroupEval without domain knowledge(L2 abilities). The following abbreviations are used: EC for Environmental Conditions; PI for Physical Infrastructure; OC for Operational Constraints; Obj for Objects; and Zone for Zones;
EU for Event Understanding; BMA for Behavior \& Maneuver Analysis; SR for Spatial Reasoning; RHA for Risk \& Hazard Assessment; CR for Causal \& Responsibility.
Additionally, the results of L1 abilities are also reported, namely, the overall performance under each category correspondingly.}
\centering
\label{tab:l2_without_dk}

\begin{tabular}{ccccccccccccc}
\toprule
\multirow{2}{*}{\textbf{VLLMs}} &
\multicolumn{6}{c}{\textbf{Perception}} & \multicolumn{6}{c}{\textbf{Reasoning}} \\
\cmidrule(lr){2-7} \cmidrule(lr){8-13}

& \textbf{Overall} & \textbf{EC} & \textbf{PI} & \textbf{OC} & \textbf{Obj} & \textbf{Zone}
& \textbf{Overall} & \textbf{EU} & \textbf{BMA} & \textbf{SR} & \textbf{RHA} & \textbf{CR}  \\
\midrule
LLaMA-VID-7B~\cite{li2025llama} & 7.8\% & 20.0\% & 5.6\% & 5.4\% & 0\% & 12.1\% 
& 6.6\% & 7.8\% & 0\% & 6.5\% & 10.2\% & 2.3\% \\
LLaMA-VID-13B~\cite{li2025llama} & 11.1\% & 20.0\% & 11.1\% & 5.4\% & 9.1\% & 12.1\% 
& 8.6\% & 5.9\% & 0\% & 10.4\% & 13.6\% & 4.5\% \\
LLaVA-One-Vision-5B~\cite{li2024llava} & 11.7\% & 36.7\% & 0\% & 5.4\% & 11.4\% & 9.1\% & 7.2\% & 0\% & 6.7\% & 2.6\% & 19.3\% & 0\% \\
Qwen2-VL-7B~\cite{wang2024qwen2} & 13.3\% & 20.0\% & 13.9\% & 5.4\% & 13.6\% & 15.2\% & 
11.7\% & 3.9\% & 3.3\% & 11.7\% & 17.0\% & 15.9\% \\
LLaVA-Next-Video-7B~\cite{zhang2024llavanextvideo} & 16.1\% & 40.0\% & 2.8\% & 13.5\% & 11.4\% & 18.2\% 
& 15.2\% & 9.8\% & 16.7\% & 10.4\% & 22.7\% & 13.6\% \\
Video-LLaVa-7B~\cite{lin2023video} & 18.9\% & 63.3\% & 5.6\% & 13.5\% & 11.4\% & 9.1\% 
& 20.7\% & 11.8\% & 20.0\% & 7.8\% & 31.8\% & 31.8\% \\
PLLaVA-7B~\cite{xu2024pllava} & 20.6\% & 50.0\% & 2.8\% & 13.5\% & 20.5\% & 21.2\% 
& 18.6\% & 11.8\% & 10.0\% & 11.7\% & 33.0\% & 15.9\% \\
LLaVA-Next-Video-34B~\cite{zhang2024llavanextvideo} & 23.9\% & 53.3\% & 22.2\% & 24.3\% & 11.4\% & 15.2\% 
& 19.0\% & 9.8\% & 13.3\% & 18.2\% & 29.5\% & 13.6\% \\
LLaVA-One-Vision-7B~\cite{li2024llava} & 26.7\% & 70.0\% & 19.4\% & 21.6\% & 13.6\% & 18.2\% 
& 20.3\% & 9.8\% & 16.7\% & 16.9\% & 36.4\% & 9.1\% \\
PLLaVA-13B~\cite{xu2024pllava} & 27.2\% & 60.0\% & 16.7\% & 21.6\% & 27.3\% & 15.2\% 
& 21.0\% & 11.8\% & 13.3\% & 16.9\% & 36.4\% & 13.6\% \\
Qwen2-VL-72B~\cite{wang2024qwen2} & 28.3\% & 53.3\% & 22.2\% & 32.4\% & 20.5\% & 18.2\% & 
30.3\% & 33.3\% & 13.3\% & 24.7\% & 42.0\% & 25.0\% \\
Qwen2-VL-2B~\cite{wang2024qwen2} & 33.3\% & 83.3\% & 30.6\% & 24.3\% & 22.7\% & 15.2\% & 
27.2\% & 17.6\% & 10.0\% & 20.8\% & 47.7\% & 20.5\% \\
MiniCPM-V~\cite{yao2024minicpm} & 35.0\% & 73.3\% & 13.9\% & 32.4\% & 31.8\% & 30.3\% 
& 28.6\% & 29.4\% & 10.0\% & 18.2\% & 48.9\% & 18.2\% \\
LLaVA-One-Vision-72B~\cite{li2024llava} & 38.3\% & 70.0\% & 19.4\% & 40.5\% & 34.1\% & 33.3\% 
& 34.5\% & 27.5\% & 26.7\% & 31.2\% & 47.7\% & 27.3\% \\
\bottomrule
\end{tabular}

\end{table*}

\subsection{Perception}
\begin{itemize}
\item {i): Environmental Conditions (EC): Focuses on external factors affecting perception.}
\begin{itemize}
    \item Atmospheric Conditions: Includes factors like Light and Weather that impact visibility.
    \item Road Surface Condition: Essential for traction and vehicle control, could be classified as "Snowy", "Dry" or "Wet".
\end{itemize}
\item{ii): Physical Infrastructure (PI): Pertains to the static aspects of the road environment.}
\begin{itemize}
    \item Road Geometry: Includes road alignment and grade affect vehicle dynamics.
    \item Lane Configuration: Includes the number of contiguous travel lanes and the number of through travel lanes related to road layout.
\end{itemize}
\item{iii): Operational Constraints (OC): Dynamic elements constraining vehicle operation.}
\begin{itemize}
    \item Traffic Conditions: Includes the factors such as the traffic flow and traffic density that influence driving decisions.
    \item Traffic Control Devices: The traffic control includes signs and signals that dictate rules.
    \item Visibility Constraints: Visual obstructions impact the driver's line of sight.
\end{itemize}
\item{iv): Objects (Obj): Identifies entities the vehicle must detect and interpret.}
\begin{itemize}
    \item Static Objects: Includes factors such as the number of objects or animals involved in an event, or the specification of other vehicles, pedestrians, animals, or objects that are involved in the event.
    \item Other Road Users: How many other motorists or non-motorists were involved in the event depicted in the video?
\end{itemize}
\item{v): Zones (Zone): Specific areas with unique driving rules or conditions.}
\begin{itemize}
    \item Geographic Zones: The description of surroundings that may influence the flow of traffic, such as  Urban, School, Open Residential, Business, or other settings.
    \item Intersection and Junctions: Includes factors such as the relation to the junction and the influence of the intersection on the vehicle's movement.
\end{itemize}
\end{itemize}

\subsection{Reasoning}
\begin{itemize}
\item {i): Event Understanding (EU): Comprehends the nature and context of events.}
\begin{itemize}
    \item Event Characteristics: Comprehends factors from video, such as the type of incident or the nature of a safety-critical event.
    \item Event Triggers: The precipitating event that initiated the sequence leading to the crash or near-crash, i.e., identifies causes of the event.
    \item Event Severity: Understand the severity of an event, if it is a crash, infer from the video what is the severity of the crash.
\end{itemize}
\item{ii):  Behavior \& Maneuver Analysis (BMA): Evaluates actions taken by the vehicle.}
\begin{itemize}
    \item Vehicle Maneuvers: Analyze the vehicle responses before, on, and after the safety-critical event, such as the pre-incident maneuver, the evasive maneuver, and the post-maneuver control.
    \item Maneuver Evaluation: The maneuver judgment assesses decision quality.
\end{itemize}
\item{iii): Spatial Reasoning (SR): Understands spatial relationships and positioning.}
\begin{itemize}
    \item Road and Terrain Features: Infer the road alignment, grade, and locality that impact navigation.
    \item Lane Positioning: Comprehends the context related to lane usage, such as which lane was the subject vehicle in during the event in the video.
    \item Spatial Relationships: Reason about the spatial relationships that help in situational awareness, for example, where was the other vehicle, pedestrian, animal, or object located in relation to the subject vehicle.
\end{itemize}
\item{iv): Risk \& Hazard Assessment (RHA): Identifies and evaluates potential dangers.}
\begin{itemize}
    \item Environmental Hazards: Understanding the environmental factors that affect safety.
    \item Traffic Hazards: Reason on the factors that might pose risks to the subject vehicle.
    \item Obstruction Hazards: Infer from the video what obstructions may require evasive actions.
\end{itemize}
\item{v): Causal \& Responsibility (CR): Determines cause and responsibility of a safety-critical event.}
\begin{itemize}
    \item Fault Analysis: Assesses liability, i.e., which driver or non-motorist (if any) committed an error that led to the event.
    \item Cause and Effect: Explain why an event occurred.
\end{itemize}
\end{itemize}

\section{Model Configurations}
In Table~\ref{tab:details_of_VLLMs}, we present a comprehensive overview of all VLLMs evaluated in DVBench. This table includes specific details on each model's language backbone, vision encoder, and their corresponding number of parameters, offering insight into the architecture and scale of each model evaluated.

\begin{figure*}[h]
  \centering
  \includegraphics[width=\textwidth, trim=120pt 120pt 120pt 120pt, clip]{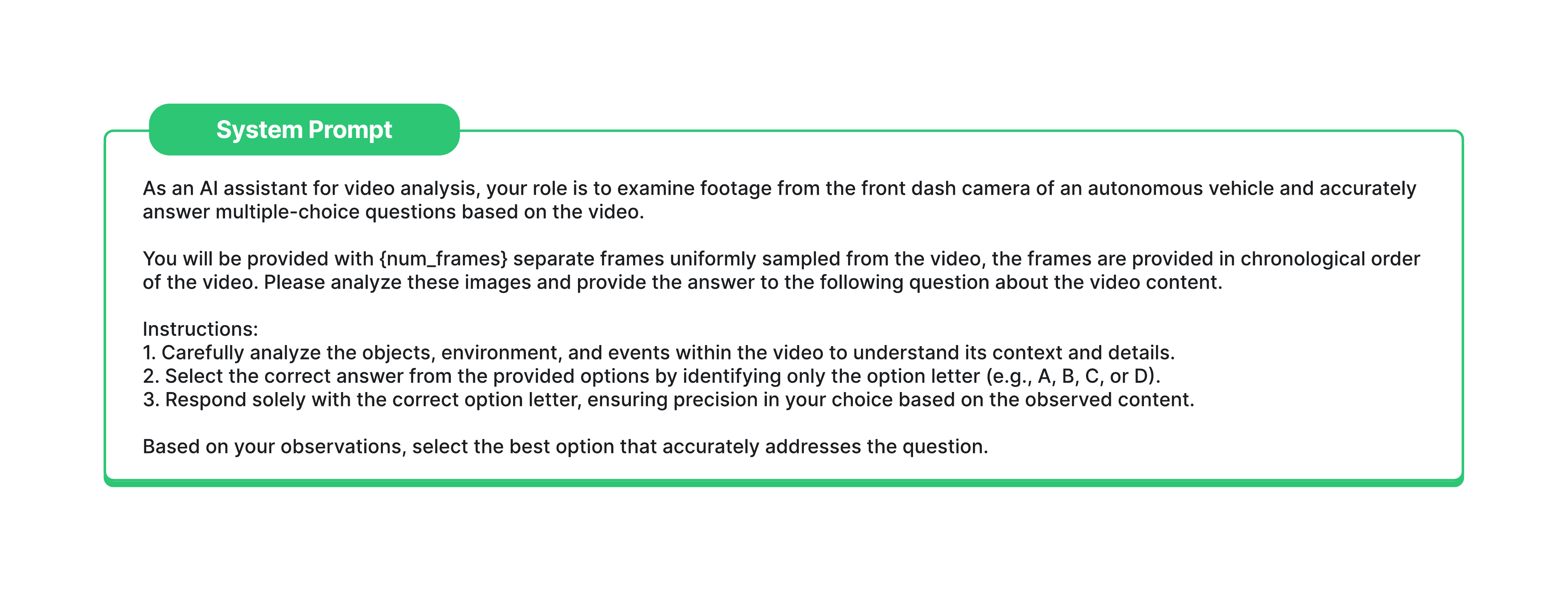}
  \caption{The system prompt for VLLM answer generation in DVBench.}
  \label{fig:system_prompt}
\end{figure*}

\begin{figure*}[h]
  \centering
  \includegraphics[width=\textwidth, trim=120pt 120pt 120pt 120pt, clip]{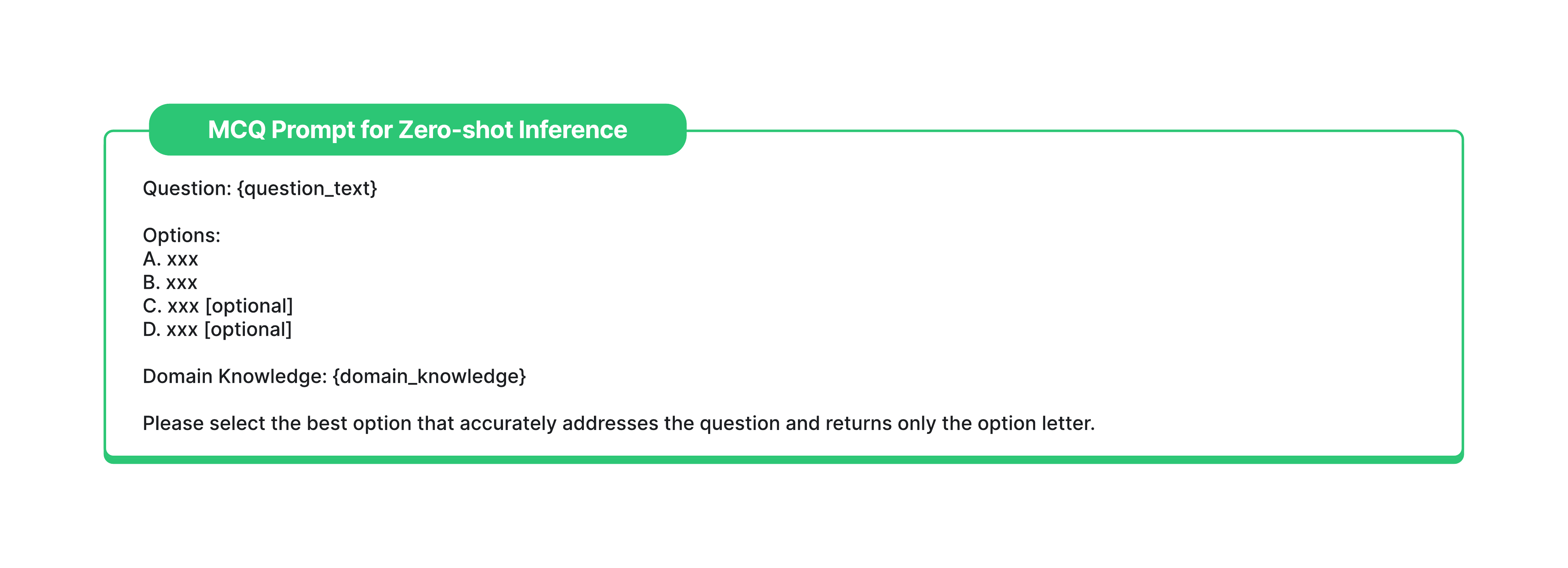}
  \caption{The prompt template for zero-shot VLLM inference.}
  \label{fig:mcq_prompt}
\end{figure*}

\section{Model Performance without Domain Knowledge}
In this section, we give detailed results about the performance of various VLLMs on DVBench without domain knowledge. Specifically, We present the evaluation results of L1 and L2 abilities in Table~\ref{tab:l2_without_dk}. Comparing these results with those obtained using domain knowledge, we observe a significant improvement in accuracy when domain knowledge is integrated. This observation aligns with the findings discussed in Section~\ref{domain knowledge matters}.

\newpage

\section{Prompt Templates}
To facilitate a fair comparison of model performance, all experiments in this paper are conducted using zero-shot inference. Figure ~\ref{fig:system_prompt} and Figure ~\ref{fig:mcq_prompt} illustrate the System Prompt and MCQ Question-Answering Prompt employed in our DVBench performance evaluation.

\section{Detailed Statistics of the MCQ Bank}
To further analyze the coverage and distribution of our Multiple-Choice Question (MCQ) bank, we provide a breakdown of the MCQ counts for each ability. The statistics are structured hierarchically, encompassing Level 1 categories (Perception and Reasoning), Level 2 subcategories (e.g., Environmental Conditions, Objects, and Zones), and Level 3 specific abilities (e.g., Atmospheric Conditions, Vehicle Maneuvers, and Traffic Hazards). 

As shown in Table~\ref{tab:category_levels}, the MCQ bank spans 25 distinct Level 3 abilities with a total of approximately 10,000 questions. Each level progressively refines the scope of assessment, enabling targeted evaluation of specific capabilities. These statistics offer a clear view of the MCQ bank's depth and its potential to systematically benchmark VLLMs' understanding of safety-critical and normal driving scenarios.

\begin{table*}[h]
\caption{The Number of Multiple-Choice Questions (MCQs) for Each Ability Across Different Levels}
\centering
\label{tab:category_levels}
\begin{tabular}{l|c|c}
\toprule
\textbf{Ability} & \textbf{Level} & \textbf{\# MCQs} \\
\midrule
Perception & Level 1 & 3925 \\
Reasoning & Level 1 & 6338 \\
\midrule
Perception/Environmental Conditions & Level 2 & 603 \\
Perception/Objects & Level 2 & 999 \\
Perception/Operational Constraints & Level 2 & 721 \\
Perception/Physical Infrastructure & Level 2 & 802 \\
Perception/Zones & Level 2 & 800 \\
Reasoning/Behavior \& Maneuver Analysis & Level 2 & 802 \\
Reasoning/Causal and Responsibility & Level 2 & 999 \\
Reasoning/Event Understanding & Level 2 & 1008 \\
Reasoning/Risk \& Hazard Assessment & Level 2 & 1924 \\
Reasoning/Spatial Reasoning & Level 2 & 1605 \\
\midrule
Perception/Environmental Conditions/Atmospheric Conditions & Level 3 & 402 \\
Perception/Environmental Conditions/Road Surface Conditions & Level 3 & 201 \\
Perception/Objects/Other Road Users & Level 3 & 602 \\
Perception/Objects/Static Objects & Level 3 & 397 \\
Perception/Operational Constraints/Traffic Conditions & Level 3 & 401 \\
Perception/Operational Constraints/Traffic Control Devices & Level 3 & 198 \\
Perception/Operational Constraints/Visibility Constraints & Level 3 & 122 \\
Perception/Physical Infrastructure/Lane Configuration & Level 3 & 400 \\
Perception/Physical Infrastructure/Road Geometry & Level 3 & 402 \\
Perception/Zones/Construction Zones & Level 3 & 199 \\
Perception/Zones/Geographic Zones & Level 3 & 200 \\
Perception/Zones/Intersection and Junctions & Level 3 & 401 \\
Reasoning/Behavior \& Maneuver Analysis/Maneuver Evaluation & Level 3 & 200 \\
Reasoning/Behavior \& Maneuver Analysis/Vehicle Maneuvers & Level 3 & 602 \\
Reasoning/Causal and Responsibility/Cause and Effect & Level 3 & 598 \\
Reasoning/Causal and Responsibility/Fault Analysis & Level 3 & 401 \\
Reasoning/Event Understanding/Event Characteristics & Level 3 & 411 \\
Reasoning/Event Understanding/Event Severity & Level 3 & 401 \\
Reasoning/Event Understanding/Event Triggers & Level 3 & 196 \\
Reasoning/Risk \& Hazard Assessment/Environmental Hazards & Level 3 & 725 \\
Reasoning/Risk \& Hazard Assessment/Obstruction Hazards & Level 3 & 600 \\
Reasoning/Risk \& Hazard Assessment/Traffic Hazards & Level 3 & 599 \\
Reasoning/Spatial Reasoning/Lane Positioning & Level 3 & 401 \\
Reasoning/Spatial Reasoning/Road and Terrain Features & Level 3 & 602 \\
Reasoning/Spatial Reasoning/Spatial Relationships & Level 3 & 602 \\
\bottomrule
\end{tabular}
\end{table*}

\end{document}